\documentclass[lettersize,journal]{IEEEtran}
\usepackage{amsmath,amsfonts}
\usepackage{array}
\usepackage[caption=false,font=normalsize,labelfont=sf,textfont=sf]{subfig}
\usepackage{textcomp}
\usepackage{stfloats}
\usepackage{url}
\usepackage{verbatim}
\usepackage{graphicx}
\usepackage{amssymb}
\usepackage{color} 
\usepackage{threeparttable}
\usepackage{amsthm,amsmath,amssymb}
\usepackage{mathrsfs}
\usepackage{indentfirst}
\usepackage{algorithm} 
\usepackage{makecell}
\usepackage{algpseudocode}
\usepackage{multirow}
\usepackage{url}
\usepackage{footnote}
\usepackage{cite}
\usepackage{array}
\usepackage{verbatim}
\usepackage{colortbl}
\usepackage{booktabs}
\usepackage{bm}
\usepackage[hidelinks,colorlinks=true,linkcolor=blue,citecolor=blue]{hyperref}

\newcommand{\tabincell}[2]{\begin{tabular}{@{}#1@{}}#2\end{tabular}}
\usepackage{balance}
\begin{document}
\title{Global Context Aggregation Network for Lightweight Saliency Detection of Surface Defects }
\author{ Feng Yan, Xiaoheng Jiang, Yang Lu, Lisha Cui, Shupan Li, Jiale Cao, \par 
Mingliang Xu, and Dacheng Tao, Fellow, IEEE
\thanks{
Manuscript created August 2023; This work was supported in part by the Nation Key Research and Development Program of China under Grant 2021YFB3301500; in part by the National Natural Science Foundation of China under Grant 62172371, U21B2037, 62102370, 62106232; in part by Natural Science Foundation of Henan Province under Grant 232300421093 and the Foundation for University Key Research of Henan Province under Grant 21A520040; in part by CAAI-Huawei MindSpore OpenFund. (\emph{Corresponding author: Xiaoheng Jiang})

Feng Yan, Xiaoheng Jiang, Yang Lu, Lisha Cui, Shupan Li, and Mingliang Xu are with School of Computer Science and Artificial Intelligence, Zhengzhou University, Zhengzhou, China; Engineering Research Center of Intelligent Swarm Systems, Ministry of Education, Zhengzhou, China; National Supercomputing Center in Zhengzhou, Zhengzhou, China (e-mail: ieyanfeng@163.com, jiangxiaoheng@zzu.edu.cn, ieylu@zzu.edu.cn, ielscui@zzu.edu.cn,iespli@zzu.edu.cn, iexumingliang@zzu.edu.cn)

Jiale Cao is with School of Electrical and Information Engineering, Tianjin University, Tianjin, China (e-mail: connor@tju.edu.cn)

Dacheng Tao is with the Sydney AI Centre and the School of Computer Science, Faculty of Engineering, The University of Sydney, Darlington, NSW 2008, Australia (e-mail: dacheng.tao@gmail.com)

}}

\markboth{IEEE Transactions on Cybernetics,~Vol.~xx, No.~xx, xx~2023}%
{How to Use the IEEEtran \LaTeX \ Templates}

\maketitle


\begin{abstract} 
Surface defect inspection is a very challenging task in which surface defects usually show weak appearances or exist under complex backgrounds. 
Most high-accuracy defect detection methods require expensive computation and storage overhead, making them less practical in some resource-constrained defect detection applications. Although some lightweight methods have achieved real-time inference speed with fewer parameters, they show poor detection accuracy in complex defect scenarios.
To this end, we develop a Global Context Aggregation Network (GCANet) for lightweight saliency detection of surface defects on the encoder-decoder structure. 
First, we introduce a novel transformer encoder on the top layer of the lightweight backbone, which captures global context information through a novel Depth-wise Self-Attention (DSA) module. The proposed DSA performs element-wise similarity in channel dimension while maintaining linear complexity.
In addition, we introduce a novel Channel Reference Attention (CRA) module before each decoder block to strengthen the representation of multi-level features in the bottom-up path.
The proposed CRA exploits the channel correlation between features at different layers to adaptively enhance feature representation.
The experimental results on three public defect datasets demonstrate that the proposed network achieves a better trade-off between accuracy and running efficiency compared with other 17 state-of-the-art methods.
Specifically, GCANet achieves competitive accuracy (91.79\% $F_{\beta}^{w}$, 93.55\% $S_\alpha$, and 97.35\% $E_\phi$) on SD-saliency-900 while running 272fps on a single gpu.

\end{abstract}


\section{Introduction}
\IEEEPARstart{S}{urface} defect inspection is an important task for industrial quality control. Manual inspection is labor-intensive, time-consuming, and low efficient. 
Traditional machine vision methods mostly depend on effective texture features \cite{hanmandlu2015detection, SONG2020LBP, Jiang2020Gabor, Berwo2021Canny}, such as statistical features, filter-based features, and model-based features. 
The manually designed features cannot detect complicated defects effectively for lack of semantic information.  
Moreover, these methods show bad reusability and generalization since they are designed for specific surface defects.

Recently, due to the strong feature representation ability of neural networks, deep learning methods have made great advances in industrial applications, such as magnetic tile\cite{huang2020surface}, road\cite{li2021fast,fan2022rethinking}, rail\cite{niu2021Unsupervised,Zhang2021MCnet}, and steel\cite{Song2020EDRNet,Zhou2022DACNet,wan2023lfrnet}. These methods mainly contain image-level\cite{bovzivc2021mixed, yang2022Semic,Gaurab2021Interleaved}, object-level\cite{su2021baf,Cui2021SDD,Ni2022ANet}, and pixel-level \cite{Zhou2022DACNet, wan2023lfrnet, Sampath2023Attention, wang2023rern} inspections.  
Although the above methods have achieved promising performance in defect detection, they mostly are involved in amounts of parameters and substantial computational overhead. 
For example, DACNet \cite{Zhou2022DACNet} has 98M parameters and 143G floating point operations (FLOPs), with a speed of 39 FPS on an NVIDIA RTX 3060 Ti, showing limitations in real-time and resource-constrained scenes. 
Although existing lightweight detection networks \cite{zhang2022fdsnet,fan2021rethinking,peng2022pp,xu2023pidnet,liu2021lightweight,liu2021samnet,li2021Depthwise} can achieve real-time detection with lower computational costs, they show poor performance in some defect scenes due to the complexity of defects. 
The main challenges of defects are as follows.
(i) \emph{weak appearance}. Defects usually show inconspicuous appearances, such as small size, thin scratches, and low contrast with the background. These weak properties make it challenging to detect complete defect regions. 
(ii) \emph{complicated background}. There are some distractions (e.g., stains, shadows, and random lighting) in the background, which may lead to false detection results. 

To improve the performance of lightweight networks in the defect detection task, two main problems should be considered. 
First, global context information is crucial for the detection of weak defects. As suggested in \cite{peng2021conformer, Xie2022PGNet}, global information is beneficial for detecting complete object regions.
However, it is difficult for lightweight CNNs to learn global dependencies because of limited receptive fields. 
To learn effective global information, Liu et al. \cite{Liu21PamiPoolNet} introduce a pyramid pooling module (PPM) \cite{zhao2017pyramid} after the final layer of CNNs. But pooling operations may damage spatial details.
Benefiting from Multi-head Self-Attention (MSA), the Transformer architecture \cite{vaswani2017attention} shows a powerful ability at modeling long-range dependencies. 
MSA models global dependencies by explicitly computing all pairwise similarities, i.e. each position is computed similarity with other all positions.
But this brings quadratic computational complexity with the spatial size, which limits its application in the lightweight network.
In addition, it is significant to refine the feature representation when defects show under complex backgrounds. 
The low-level features contain abundant background interference compared to the high-level features. 
So simply fusing original low-level features with other features by addition or concatenation may result in some incomplete or false detection results.

To solve the aforesaid issues, we propose a Global Context Aggregation Network (GCANet) for lightweight saliency detection of surface defects. 
First, we introduce a novel transformer encoder on top of the encoder, which takes the top-level features as inputs to learn global information. To mitigate the computational overhead of MSA, we propose a novel Depth-wise Self-Attention (DSA) module with linear complexity. DSA can learn global information as MSA does. Differently, DSA produces attention weights via element-wise interaction.
Meanwhile, considering the complementary of global and local features, global features are injected into all subsequent decoding stages through shortcuts to activate more complete defect regions.
Secondly, we introduce a Channel Reference Attention (CRA) module to strengthen the feature representation. The high-level and global features are beneficial to suppress background noise because they contain richer semantic information than low-level features. CRA produces an attention map by computing the channel similarity between them and low-level features. The attention map can adaptively highlight useful defect information along the channel dimension, which enables the network to focus more on defect details.

In summary, the main contributions are as follows:
\begin{enumerate}
\itemsep=0pt
\item We propose a Global Context Aggregation Network (GCANet) for lightweight saliency detection of surface defects, which achieves fast and accurate defect detection.
\item We present a novel transformer block based on the proposed Depth-wise Self-Attention (DSA). The DSA can learn global information with linear complexity.
\item We present a Channel Reference Attention (CRA) module to refine the expression of features. The CRA selectively emphasizes meaningful defect detail features by learning similarities between cross-level features.
\item Extensive experiments on three public defect datasets demonstrate the proposed model achieves the trade-off between accuracy and efficiency in defect detection scenes. 
\end{enumerate}

\section{Related Works}
Defect detection methods are roughly grouped into traditional machine vision based \cite{hanmandlu2015detection, SONG2020LBP, Jiang2020Gabor, Berwo2021Canny} and deep learning based methods \cite{bovzivc2021mixed, yang2022Semic,Gaurab2021Interleaved,su2021baf,Cui2021SDD,Ni2022ANet, wang2023rern,Zhou2022DACNet, Sampath2023Attention, wan2023lfrnet,Zhang2021MCnet,li2021fast}. Traditional machine vision methods show limitations in defect scenes, such as low accuracy and poor reusability. Here, we briefly review deep learning based pixel-wise defect detection methods and some lightweight networks.
\subsection{Deep Learning Based Pixel-wise Defect Detection Methods}
Recently, deep learning methods have made remarkable advances in the surface defect detection task due to the strong feature extraction function of neural networks. Different from image-level \cite{bovzivc2021mixed, yang2022Semic,Gaurab2021Interleaved} and object-level \cite{su2021baf,Cui2021SDD,Ni2022ANet} detection methods, pixel-level detection methods \cite{wang2023rern,Zhou2022DACNet, Sampath2023Attention, wan2023lfrnet,Zhang2021MCnet,li2021fast} obtain fine-grained detection results.
The utilization of context information is crucial for pixel-wise defect detection.
Considering the complexity of defects, Wang et al. \cite{wang2023rern} exploited channel and spatial global dependencies to strengthen the representation of defect features.
Similarly, Zhou et al. \cite{Zhou2022DACNet} deployed three convolutional branches with different depths in the encoder to learn multi-scale context information and introduce a dense attention mechanism in the decoder.
In addition, Sampath et al. \cite{Sampath2023Attention} added channel and spatial attention module in the encoder to highlight defect features and filter out background interference.
Wan et al. \cite{wan2023lfrnet} integrated effective context semantics, spatial details, and edge features to achieve accurate defect detection.
Aimed at the problem of data imbalance between defect pixels and non-defect pixels in defect images, Li et al. \cite{li2021fast} propose novel adaptive weighted cross-entropy (WCE) loss functions to train the network, which can make the network learn more information of defects.

Although the above methods achieve excellent performance in defect detection, these methods are limited in resource-constrained and real-time scenes because of substantial parameters and expensive computational overhead.
To achieve real-time defect detection, Huang et al. \cite{Huang2020Compact} constructed a compact segmentation network consisting of a lightweight encoder and decoder.
Zhang et al. \cite{zhang2022fdsnet} developed a real-time surface defect segmentation network, called FDSNet, which adopts two branches to encode edge details and semantic information of defects, respectively.

\subsection{Lightweight Networks}
Lightweight networks can achieve real-time inference because of low computational costs, i.e. fewer parameters and FOLPs. Currently, there are three mainstreams in the design of lightweight methods.
(i) \emph{lightweight backbone}. Considering the importance of multi-scale contexts, Fan et al. \cite{fan2021rethinking} presented a Short-Term Dense Concatenate (STDC) module that can capture multi-scale contexts and use it to develop a lightweight network named STDCNet.
Similarly, Liu et al. \cite{liu2021samnet} utilized the designed stereoscopically attentive multi-scale (SAM) unit to develop a lightweight SOD model called SAMNet, with fewer parameters and FLOPs compared to STDCNet.
(ii) \emph{multi-branch architecture}. Poudel et al. \cite{Poudel2019FastSCNNFS} designed a two-branch network for real-time segmentation, where context and detail branches share initial several layers to reduce computational costs.
Xu et al. \cite{xu2023pidnet} introduced three different branches to learn details, semantics, and boundary features, respectively.
(iii) \emph{lightweight module design}. 
Peng et al. \cite{peng2022pp} adopted STDCNet as the encoder and designed a unified attention fusion module to integrate low-level and high-level features.
Li \cite{li2022CorrNet} et al. introduced correlation-guided feature fusion and lightweight feature refinement block in the decoder to improve performance. 

Different from the previous works, the proposed method introduces a novel transformer block in the lightweight backbone to capture global information, considering the limited receptive field of CNNs. Each transformer block uses the proposed Depth-wise Self-Attention to learn global context information. Furthermore, the Channel Reference Attention module is presented to strengthen feature representations.

\begin{figure*}[tbh]
	\centering
	\includegraphics[scale=0.68]{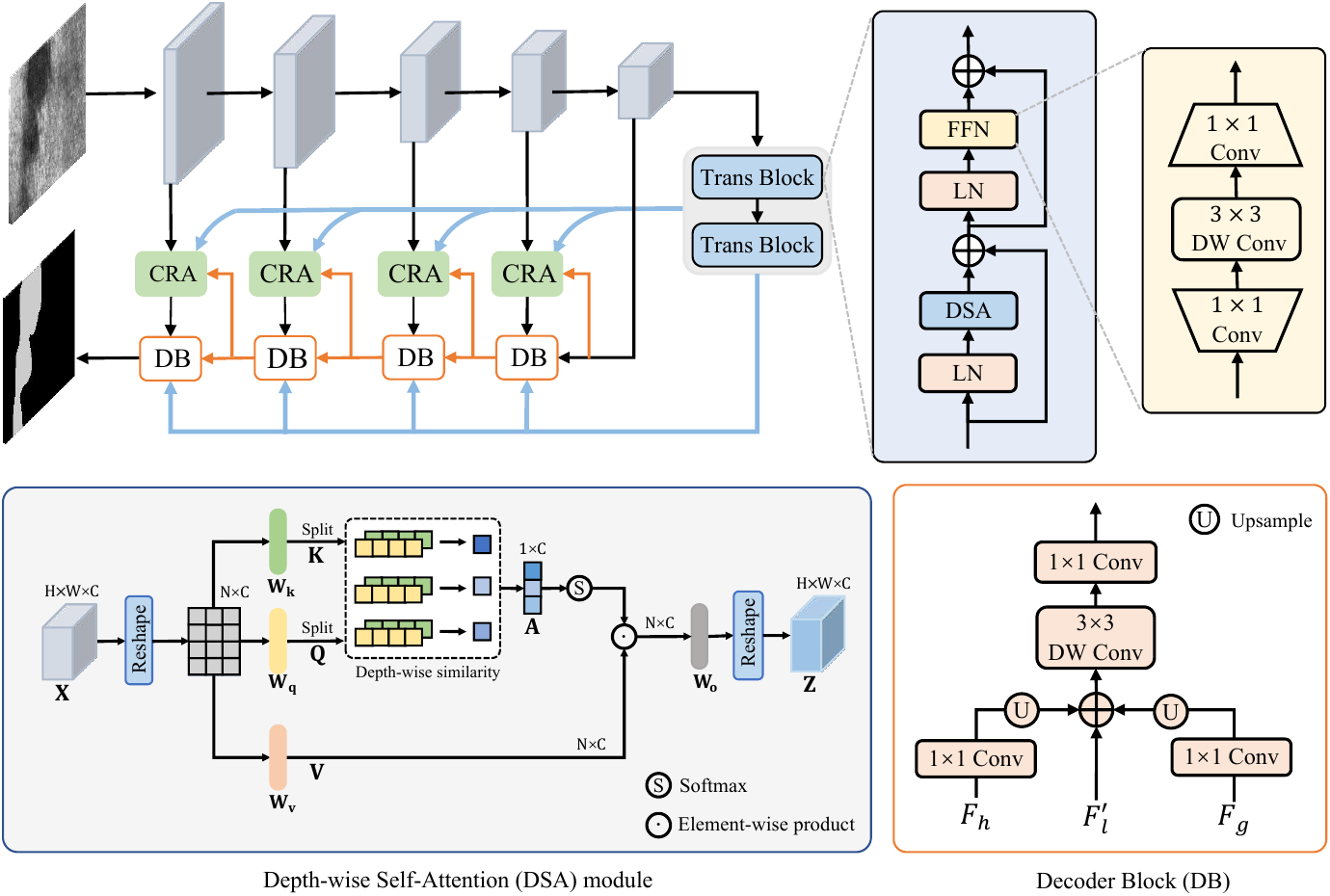}
	\caption{ The overview of the proposed GCANet. It adopts the lightweight backbone as the encoder. Following the encoder, an extra transformer encoder is introduced to capture global context information. Depth-wise Self-Attention (DSA) module can model global dependencies with linear complexity. Meanwhile, the Channel Reference Attention (CRA) module is introduced to strengthen the representation ability of features through channel interaction of cross-level features. }
	\label{fig1}
\end{figure*}

\section{Method}
In this section, we first describe the overall architecture of the proposed method in Section III-A. Then, we introduce the designed DSA and CRA modules in Section III-B and Section III-C, respectively. Finally, we present the loss function in Section III-D.

\subsection{Overall Architecture}
The proposed network is an encoder-decoder network and the details are described in Fig.\ref{fig1}. 
We adopt the backbone of lightweight network SAMNet \cite{liu2021samnet} as the encoder.
Differently, we use a depthwise separable 3$\times$3 convolution (DSConv3$\times$3) with stride 1 instead of stride 2 in the first encoding stage.
This enables the network to encode more defect details at the early layers.
With an input image $I\in \mathbb{R}^{H \times W \times 3}$ given, the dimension of output feature at encoding stage $i$ is $ \frac{H}{2^{i-1}}\times\frac{W}{2^{i-1}}\times C_i$, $C_i \in \{16,32,64,96,128\}$. 
We introduce an additional transformer encoder to learn global context semantics, which takes as input the output feature of the fifth encoding stage.  
The transformer block contains a depth-wise self-attention module (DSA) and a feed-forward network (FFN). The DSA can capture global dependencies while maintaining linear complexity.
To remedy the problem of feature dilution in the top-down path, global features are injected into subsequent each decoder block (DB) through skip connections.
The integration of global and local features is beneficial to detect complete defect objects.
Besides, we introduce a Channel Reference Attention (CRA) module before feature fusion for feature enhancement. The attention weights are dynamically computed by channel similarities between cross-layer features, which can adaptively highlight important defect features.

In each decoder block, global, high-level, and refined low-level features are aggregated together by element-wise summation and the dilated DSConv3$\times$3. The output features of decoder blocks and the transformer encoder are respectively fed into the convolution layer to produce side-output saliency predictions for deep supervision.

\subsection{Depth-wise Self-Attention} 

Transformers capture global context dependencies through multi-head self-attention (MSA). MSA computes the pairwise similarity among all spatial elements, producing an attention map with a size of $N\times N$,  where $N$ denotes the spatial size of features. This brings high computational complexity and memory usage. To mitigate this problem, we present a novel Depth-wise Self-Attention, which calculates self-attention in the channel dimension and implicitly models the global context information, as described in Fig. \ref{fig1}

Suppose $ \mathbf{X}\in \mathbb{R}^{ H \times W \times C}$ as input feature map, which is first reshaped to $\mathbf{X^\prime}\in \mathbb{R}^{N \times C}$, where $ N =H \times W$. 
$\mathbf{X^\prime}$ generates query $\mathbf{Q}$, key $\mathbf{K}$ and value $\mathbf{V}$ through different linear projections. Formally, we have
\begin{equation}
\mathbf{Q=X^\prime W_q, K=X^\prime W_k, V=X^\prime W_v}
\end{equation}
where $\mathbf{W_q}$, $\mathbf{W_k}$, and $ \mathbf{W_v}\in \mathbb{R}^{C \times C}$ represent learnable weight matrices.

DSA calculates depth-wise similarities between $\mathbf{K}$ and $\mathbf{Q}$, producing a global attention map with a size of $\mathbb{R}^{1\times C}$ instead of $\mathbb{R}^{N\times N}$. Specifically, with query $\mathbf{Q}$ and key $\mathbf{K}$ obtained, DSA splits them into $C$ key vectors $k_i$ and $C$ query vectors $q_i$ along the channel dimension, respectively, where $q_i$ and $k_i\in \mathbb{R}^{N \times 1}$. For each $q_i$ and $k_i$, DSA computes their normalized inter product as the global context descriptor $g_i \in \mathbb{R}^{1}$, which is formulated as follows:
\begin{equation}
g_i= \phi (q_i)^T \phi (k_i)
\end{equation}
where $\phi (\cdot)$ denotes the $\ell{_2}$ normalization function, which can restrict the inter-product results of $q_i$ and $k_i$ in the range of $[-1, 1]$. 

The obtained global context descriptors $g_1, ..., g_C$ are concatenated together and multiplied by a learnable scale factor $\alpha$, producing a global context attention map $\mathbf{A}\in \mathbb{R}^{1\times C}$ through a Softmax function. The \textbf{A} is expanded as $\mathbb{R}^{N \times C}$ along the spatial dimension, performing an element-wise multiplication with \textbf{V}. Mathematically, we have:

\begin{equation}
\mathbf{Z} = Softmax(\alpha \mathop{Concat}\limits_{i=1}^{C}(g_i))\odot \mathbf{V}
\end{equation}
where $\odot$ denotes the element-wise product. The output feature is linearly projected again via a learnable weight matric $ \mathbf{W_o}\in \mathbb{R}^{C \times C}$ to produce the final result.

DSA generates a global context vector with the same channel number as input features, and weights features in the channel dimension. So it maintains linear computational complexity, computed as:
\begin{equation}
\mathcal{O}(DSA)=4C^2N+2CN
\end{equation}

In summary, DSA implicitly models global dependencies through element-wise interaction between features, which brings linear complexity.

\subsection{Channel Reference Attention module} 
Considering that low-level features contain abundant background noise except for important defect details, we design a Channel Reference Attention module (CRA) to strengthen the representation of features in the encoder. The high-level and global features contain richer context semantics than low-level features, which is beneficial for suppressing background interference. CRA exploits correlations between cross-level features in the channel dimension to focus on meaningful defect details, boosting the representation ability of low-level features.

\begin{figure}[t]
	\centering
	\includegraphics[scale=0.42]{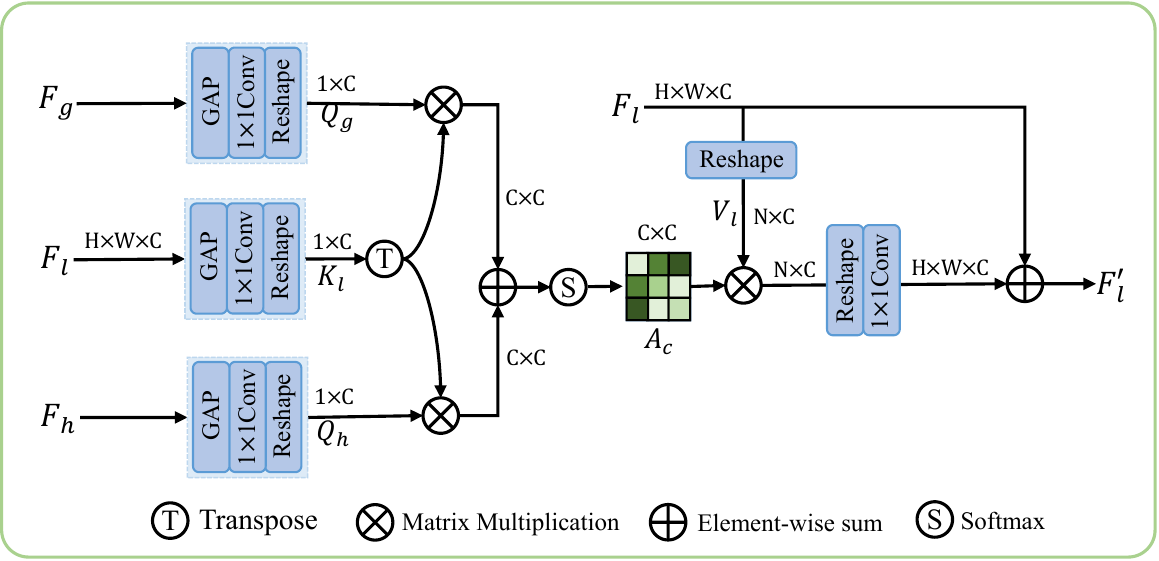}
	\caption{Illustration of the Channel Reference Attention (CRA) module.}
	\label{fig2}
\end{figure}

As described in Fig. \ref{fig2}, CRA is fed into low-level features, high-level features, and global features, which are denoted as $F_l$, $F_h$, and $F_g$, respectively. Firstly, $F_l$, $F_h$ and $F_g$ are projected to keys $K_l$, queries $Q_h$ and $Q_g$ through global average pooling (GAP) operation, $1\times1$ convolution, and ReLU function, respectively. Mathematically, we have:
\begin{equation}
K_l= \delta(Conv_{1\times1}(GAP(F_l)))
\end{equation}
\begin{equation}
Q_h= \delta(Conv_{1\times1}(GAP(F_h)))
\end{equation}
\begin{equation}
Q_g= \delta(Conv_{1\times1}(GAP(F_g)))
\end{equation}
where $\delta$ denotes the ReLU function, $Conv_{1\times1}$ denotes the $1\times1$ convolution.

Next, $K_l$, $Q_h$ and $Q_g$ are reshaped into $\mathbb{R}^{1 \times C}$ from $\mathbb{R}^{1 \times 1 \times C}$. The transpose of the reshaped $Q_h$ and $Q_g$ are multiplied by the reshaped $K_l$, respectively, generating channel similarity matrix $A_1$ and $A_2 \in \mathbb{R}^{C\times C}$. The $A_1$ and $A_2$ are added together and multiplied by a learnable temperature parameter $\tau$, generating channel attention weights $A_c$ through Softmax function. Mathematically, we have:

\begin{equation}
A_1=Q_h^TK_l, \quad A_2=Q_g^TK_l
\end{equation}

\begin{equation}
A_c=Softmax(\frac{A_1+A_2}{\tau})
\end{equation}
where $\tau$ is used to adaptively adjust the contribution of attention weights.

Finally, with $A_c$ obtained, we reshape $F_l$ into $\mathbb{R}^{ N\times C}$ and perform matrix multiplication with $A_c$, where $N=H \times W$. The outputs are reshaped to $\mathbb{R}^{ H\times W\times C}$ and projected through a $1\times 1$ convolution again. Meantime, we add a residual connection between the output and the original $F_l$. Mathematically, we have:
\begin{equation}
F_l^{\prime}=Conv_{1\times1}(F_lA_c)+F_l
\end{equation}

Similar to the MSA, we split the number of channels into $h$ heads to jointly focus on important defect details from different subspaces. Specifically, we employ $h=4$ for each CRA.

\subsection{Loss Function}
We adopt deep supervision strategy\cite{Hou2017deeply} and hybrid loss function to train network. Mathematically, the loss function is computed as:
\begin{equation}
\left\{\begin{array}{ll}
\mathcal{L}_{\text {total }}=\sum\limits_{i=1}^{5}\mathcal{L}({\rm \bf G}, {\rm \bf S}_i)\\
\mathcal{L}({\rm \bf G}, {\rm \bf S}_i) =\mathcal{L}_{bce}+\mathcal{L}_{iou}+\mathcal{L}_{ssim}
\end{array}\right.
\end{equation}
where ${\rm \bf S}_i$ represents the $i^{th}$ side-output saliency prediction, and \textbf{G} represents the corresponding ground-truth. $\mathcal{L}_{bce}$, $\mathcal{L}_{iou}$, and $\mathcal{L}_{ssim}$ represents the binary cross-entropy (BCE) loss \cite{BoerKMR05}, intersection over union (IoU) loss \cite{Rahman2016Optimizing} and structural similarity (SSIM) loss \cite{Wang2003structural}, respectively.

BCE loss is a pixel-level classification loss, defined as:
\begin{equation}
\begin{aligned}
\mathcal{L}_{bce}=-\frac{1}{HW}\sum_{x=1}^{H}\sum_{y=1}^{W}{[G_{xy}log(S_{xy})
+ \widetilde{G}_{xy}log(\widetilde{S}_{xy})]}
\end{aligned}
\end{equation}
where $H$ and $W$ represent the height and width of \textbf{G}, respectively. $G_{xy}$ and $S_{xy}$ represent the label and prediction of \textbf{G} and \textbf{S} at position $(x,y)$, respectively. And $\widetilde{S}_{xy}=1-S_{xy}$, $\widetilde{G}_{xy}=1-G_{xy}$. 

IoU loss is computed based on the IoU measure, which is beneficial for the model to focus on defect pixels, defined as:
\begin{equation}
\begin {aligned} 
\mathcal{L}_{iou}=1-\frac{\sum\limits_{x=1}^{H}\sum\limits_{y=1}^{W}{G_{xy}}\cdot S_{xy}}{\sum\limits_{x=1}^{H}\sum\limits_{y=1}^{W}{(G_{xy}+S_{xy}-G_{xy}}\cdot S_{xy})}
\end{aligned} 
\end{equation}

SSIM loss is computed based on the structural similarity measure, defined as:
\begin{equation}
 \mathcal{L}_{ssim}=1-\frac{(2 \mu_{a} \mu_{b}+\xi_{1})(2 \sigma_{ab}+\xi_{2})}{(\mu_{a}^{2}+\mu_{b}^{2}+\xi_{1})(\sigma_{a}^{2}+\sigma_{b}^{2}+\xi_{2})}
\end{equation}
where $a$ and $b$ denote two $k\times k$ patches cropped from \textbf{S} and \textbf{G}, respectively.
$\mu_a$, $\mu_b$, $\sigma_a$, $\sigma_b$, and $\sigma_{ab}$ represent means, standard deviations and covariance of patches $a$ and $b$, respectively. $\xi_1 = 0.01^2$ and $\xi_2= 0.03^2$.
\section{Experiments}
\subsection{Datasets}
To validate the effectiveness of the proposed GCANet for surface defect detection, we conduct experiments on the following surface defect datasets.

\emph{1) SD-saliency-900}\cite{Song2020EDRNet}: There are three typical types of strip steel defects: inclusions, patches, and scratches. Each defect sample has a resolution of 200$\times$200 with a corresponding pixel-level label. These defects are characterized by low contrast, various types, different scales, and cluttered ebackground, which bring difficulties to accurate detection. Each category of defects includes 300 images. We adopt the same training dataset (810 images) and test dataset (900 images) as previous works \cite{Song2020EDRNet, Zhou2022DACNet, wang2023defect} in the experiment. 

\emph{2) Magnetic tile}\cite{huang2020surface}: There are 392 defect images, including five categories of defects: uneven, fray, crack, blowhole, and break. Each defect image has the corresponding fine-grained pixel-level label. 
Defects show various types and scales, complicated background noise (e.g.stains and shallow), and low contrast.
In the experiment, the dataset includes 194 training images and 198 test images, which are obtained by randomly dividing images of each category at a ratio of 1:1.

\emph{3) DAGM 2007}\cite{wieler2007weakly}: There are 10 types of defects with each generated by a specific defect model and texture model. Each defect image contains a defect object, roughly labeled with an ellipse. Defects with various types and complex backgrounds bring challenges to detection. 
The dataset includes 1046 training images and 1054 test images. In the experiment, the training dataset is increased to 3138 images by flipping horizontally and vertically.

\subsection{Implementation Details}
The proposed method is implemented with Pytorch. 
The parameters of the encoder are initialized with the pre-trained SAMNet backbone on ImageNet. 
The experiments are conducted on a computer with NVIDIA RTX 3060 Ti.  
The network is trained with Adam optimizer, where the learning rate is set to 5e-4. 
It is trained for 900 epochs with a batch size of 8 on SD-saliency-900, 900 epochs with a batch size of 5 on Magnetic tile, and 270 epochs with a batch size of 8 on DAGM 2007, respectively.
During the training stage, the original image is resized to 256$\times$256 and randomly cropped to 224$\times$224 as the input of the network, following previous works \cite{Song2020EDRNet, Zhou2022DACNet}. 
During the test stage, the proposed network takes 256$\times$256 resolution as the input. 
In addition, we only calculate the output saliency map of the final decoding stage.
The obtained saliency prediction is sampled to the same resolution as the original image for evaluation.
 
\subsection{Evaluation Metrics}
We quantitatively evaluate the performance of various models on the following metrics.

\textbf{\emph{Precision-Recall}} (\emph{PR}) curve \cite{borji2019salient} is plotted with different precision-recall pairs with each calculated on binarized saliency map \textbf{S}. The binarized \textbf{S} are obtained by using 255 different thresholds in the range of $[0, 1]$, respectively.

\textbf{\emph{F-measure}} ($F_{\beta}$) \cite{achanta2009frequency} is a comprehensive metric that accounts for both accuracy and recall. $F_{\beta}$ of each precision-recall pair is calculated as follows: 
\begin{equation}
 F_{\beta}=\frac{(1+\beta^{2}) \text {Precision} \times \text {Recall}}{\beta^{2}\times \text {Precision}+\text {Recall}}
\end{equation}
where $\beta^2 = 0.3$ in the experiment. The \emph{F-measure} curve is plotted with calculated $F_{\beta}$ scores using precision-recall pairs under different thresholds.

\textbf{\emph{Mean Absolute Error}} (\emph{MAE})\cite{perazzi2012saliency} evaluates the difference of \textbf{S} and \textbf{G} by computing their average pixel-level absolute error, computed as:
\begin{equation}
 MAE = \frac{1}{H \times W} \sum\limits_{x=1}^H\sum\limits_{y=1}^W|S_{xy}- G_{xy}|
\end{equation}

\textbf{\emph{Weighted F-measure }}($F_{\beta}^{w}$) \cite{margolin2014evaluate} defines the generalized $F_{\beta}$ by assigning different weights to errors at different locations, computed as:
\begin{equation}
 F_{\beta}^{w}= \frac{\left(1+\beta^{2}\right)\text {Precision}^{w} \times \text {Recall}^{w}}{\beta^{2}\times \text {Precision}^{w}+\text {Recall}^{w}}
\end{equation}
where $\beta^2=1$ in the experiment.

\textbf{\emph{Structural similarity measure }}($S_{\alpha}$)\cite{fan2017structure} evaluates the structural similarity of \textbf{S} and \textbf{G} through object-aware ($S_{o}$) and region-aware ($S_{r}$) measures, defined as: 
\begin{equation}
 S_{\alpha}=\alpha * S_{o}+(1-\alpha) * S_{r}
\end{equation}
where $\alpha = 0.5$ in the experiment.

\textbf{\emph{Enhanced-alignment measure}} ($E_{\phi}$) \cite{EM} considers local pixel-level and global image-level properties, defined as:
\begin{equation}
 E_{\phi}=\frac{1}{W \times H} \sum_{(x,y)} \phi(x,y)
\end{equation}
where $\phi(x,y)$ represents the enhanced alignment matrix. We use mean $E_{\phi}$ in the experiment.

%

\begin{table*}[!t]
\renewcommand\arraystretch{1.3}
\centering

\caption{Quantitative comparisons of different methods on SD-saliency-900, Magnetic tile, and DAGM 2007. The No.1$\sim$No.5 represent defect detection models. The No.6$\sim$No.10 represent efficient models. And the No.11$\sim$No.17 represent lightweight models. "$\downarrow$" indicates that the lower the value, the better the performance, and "$\uparrow$" is the opposite.}
\label{tab1}

\setlength{\tabcolsep}{1mm}{

\begin{tabular}{c|r|r|r|r|cccc|cccc|cccc}		
\hline
\multirow{2}*{No.}&\multirow{2}*{Methods} &\multirow{2}*{\tabincell{c}{\#Param\\(M)}}&\multirow{2}*{\tabincell{c}{FLOPs\\(G)}}&\multirow{2}*{\tabincell{c}{Speed\\(FPS)}}&\multicolumn{4}{c|}{\tabincell{c}{SD-saliency-900}} & \multicolumn{4}{c|}{Magnetic tile} & \multicolumn{4}{c}{DAGM 2007} \\
\cline{6-17}
~&~&~&~&~&$\text{MAE}\downarrow$ & $F_{\beta}^{w}\uparrow$ & $S_{\alpha}\uparrow$ & $E_{\phi}\uparrow$& $\text{MAE}\downarrow$ & $F_{\beta}^{w}\uparrow$ & $S_{\alpha}\uparrow$  & $E_{\phi}\uparrow$ & $\text{MAE}\downarrow$ & $F_{\beta}^{w}\uparrow$ & $S_{\alpha}\uparrow$ &$E_{\phi}\uparrow$ \\
\hline
1&MCNet\cite{Zhang2021MCnet} 
&38.41&36.88&29.1
&0.0149& 0.9066&0.9257&0.9691
& 0.0176& 0.7537&0.8391&0.8934
&0.0046& 0.8524&0.9083&0.9556
\\
2&EDRNet\cite{Song2020EDRNet} 
&39.31&42.14&35
& 0.0130& 0.9225&0.9375&0.9754
&0.0204& 0.7828&0.8601&0.9051
&0.0046 & 0.8612 &0.9097 & 0.9550
\\
3&DACNet\cite{Zhou2022DACNet}
&98.39&142.71&39
&0.0118& 0.9275&0.9417&0.9773
&0.0194& 0.7993&0.8676&0.9249
&0.0065& 0.8227&0.8914 &0.9449
\\
4&LWNet\cite{Huang2020Compact}
&2.21&0.46&331.5
&0.0222& 0.8516&0.8917&0.9533
&0.0232& 0.6736&0.8010&0.8708
&0.0071& 0.7629&0.8644 & 0.9199
\\
5&FDSNet\cite{zhang2022fdsnet}
&0.96&0.26&340.8
&0.0185& 0.8870&0.9113&0.9648
& 0.0202& 0.7197&0.8224&0.8814
&0.0061& 0.8095&0.8817&0.9351
\\
\hline
6&BASNet\cite{Qin2019basnet}
&87.06&127.40&62.2
&0.0152& 0.9092&0.9276 & 0.9691
&0.0174& 0.8086 &0.8744& 0.9308
&0.0048 & 0.8687&0.9167 & 0.9594
\\
7&FFRNet\cite{Wang2022Focus}
&58.56&72.32&41.4
& 0.0136 & 0.9192 & 0.9353 & 0.9731
& 0.0186 & 0.7929 & 0.8577 & 0.9259
& 0.0054 & 0.8328 & 0.8908 & 0.9475

\\
8&ACCoNet\cite{li2022adjacent}
&127.01&51.26&42.1
& 0.0155 & 0.9044 & 0.9217 & 0.9668
& 0.0173 & 0.8019 & 0.8720 & 0.9178
& 0.0061 & 0.8325 & 0.8976 & 0.9498
\\

9&PGNet \cite{Xie2022PGNet} 
&72.62&18.37&33
&0.0150& 0.9087&0.9296&0.9704
&0.0174& 0.7894&0.8632&0.9143
&0.0056 & 0.8345 & 0.8999 & 0.9487
\\
10&UCTransNet \cite{2022wangUCTransNet}
&66.22&32.87&49
&0.0140& 0.9173&0.9352&0.9728
&0.0185&0.7941 &0.8683&0.9185
&0.0054& 0.8366&0.8938&0.9450
\\
\hline
11&HVPNet\cite{liu2021lightweight} 
&1.24&0.65&293.7
& 0.0173 & 0.8886 & 0.9189 &0.9599
& 0.0179 & 0.7257 & 0.8476 & 0.8698
& 0.0060& 0.8018&0.8944&0.9395
\\
12&STDC2-Seg\cite{fan2021rethinking} 
&22.30&9.16&327.8
&0.0158& 0.9007&0.9208&0.9693
&0.0206& 0.7210&0.8272&0.8856
&0.0066& 0.8186&0.8911& 0.9482
\\
13&SAMNet\cite{liu2021samnet}
&1.33&0.31&307.7
&0.0181& 0.8856&0.9181&0.9594
& 0.0180& 0.7513&0.8636&0.8935
& 0.0054& 0.8149&0.8999&0.9402
\\
14&AttaNet\cite{Song2021AttaNet} 
&12.78&2.95&300.5
&0.0165& 0.8932&0.9189&0.9688
&0.0179& 0.7613&0.8461&0.9142
&0.0077 & 0.8050 & 0.8827 & 0.9411
\\
15&PP-LiteSeg\cite{peng2022pp} 
&13.47&4.42&395.4
&0.0155& 0.9021&0.9244&0.9703
&0.0195& 0.7530&0.8425&0.9013
&0.0059& 0.8233&0.8912&0.9392
\\
16&CorrNet\cite{li2022CorrNet}
&4.08&21.08&213
& 0.0151& 0.9107&0.9272&0.9683
&0.0176&0.7913&0.8630&0.9113
&0.0075&0.8279&0.8965&0.9456\\
17&PIDNet-S \cite{xu2023pidnet}&7.57&6.30&228.5&0.0143&0.9089&0.9309&0.9654&0.0250&0.7494&0.8236&0.8839&0.0052&0.8548&0.8911&0.9529\\
\hline
18&GCANet(Ours)
&1.84&1.45&272.2
&0.0135& 0.9179&0.9355&0.9735
&0.0168& 0.8064&0.8770&0.9279
&0.0044&0.8653&0.9151&0.9593
\\
\hline
\end{tabular}}
\end{table*}

\subsection{Comparisons with State-of-the-arts}
To prove the advantage of the proposed GCANet, it is compared with 17 state-of-the-art segmentation models over three defect datasets, including defect detection models (i.e. MCNet\cite{Huang2020Compact}, EDRNet\cite{Song2020EDRNet}, DACNet\cite{Zhou2022DACNet},  FDSNet\cite{zhang2022fdsnet}, LWNet\cite{Huang2020Compact}), efficient models (i.e. BASNet\cite{Qin2019basnet}, FFRNet\cite{Wang2022Focus}, ACCoNet\cite{li2022adjacent}, PGNet \cite{Xie2022PGNet}, UCTransNet\cite{2022wangUCTransNet} ), lightweight models (i.e. HVPNet\cite{liu2021lightweight}, STDC2-Seg\cite{fan2021rethinking},  SAMNet\cite{liu2021samnet}, AttaNet\cite{Song2021AttaNet},  PP-LiteSeg\cite{peng2022pp}, CorrNet\cite{li2022CorrNet}, PIDNet-S\cite{xu2023pidnet}).
It is noted that we generate single-channel predictions at the output layer of some semantic segmentation models. (i.e.\cite{zhang2022fdsnet, fan2021rethinking, Song2021AttaNet, peng2022pp, xu2023pidnet}). And we employ the same loss function and training strategy as the proposed model for these models during the training process. 

\begin{figure}[t]
	\centering
		\includegraphics[scale=0.44]{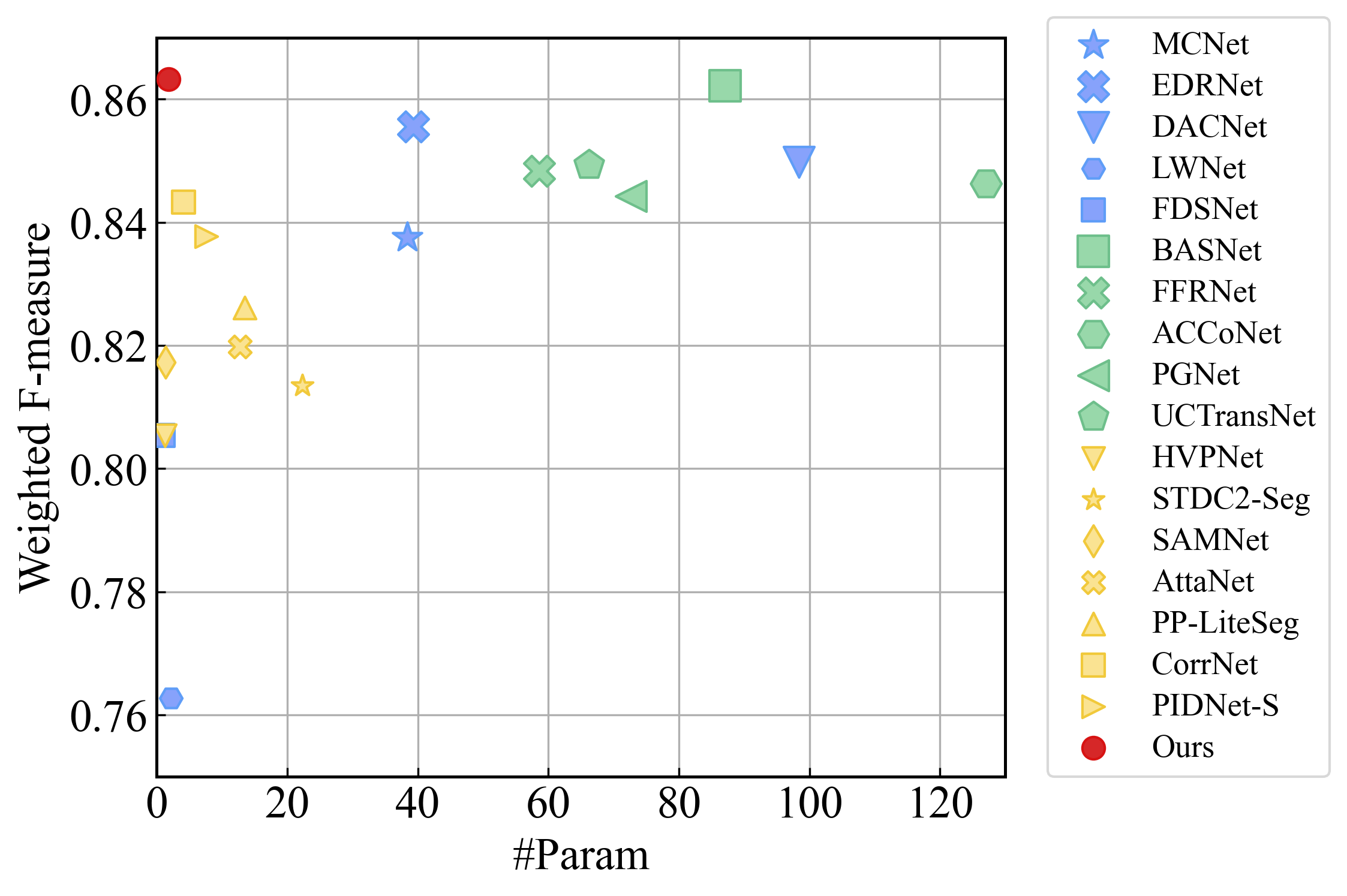}
	\caption{Illustration of the trade-off between accuracy and efficiency for different methods. The weighted F-measure ($F_{\beta}^{w}$) is the average of that on three defect datasets. }
	\label{fig3}
\end{figure}

\begin{figure*}[tbh]
	\centering
	\includegraphics[scale=0.52]{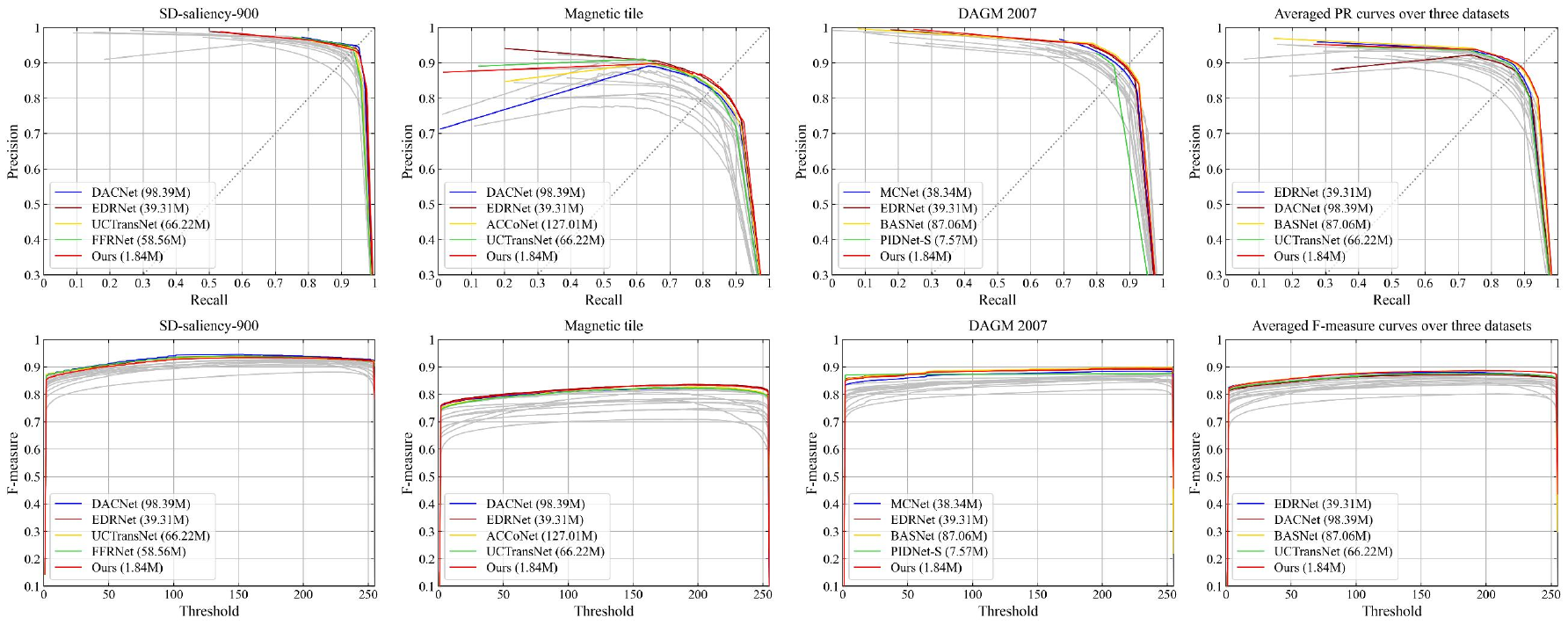}
	\caption{Comparisons of various methods in terms of PR curves and F-measure curves on different defect datasets, respectively. Note that the closer the PR curve is to the coordinates (1,1), the higher the F-measure curve, the better the performance. The top five methods are highlighted in different colors. }
	\label{fig4}
\end{figure*}

\begin{figure*}[t]
	\centering
		\includegraphics[scale=0.52]{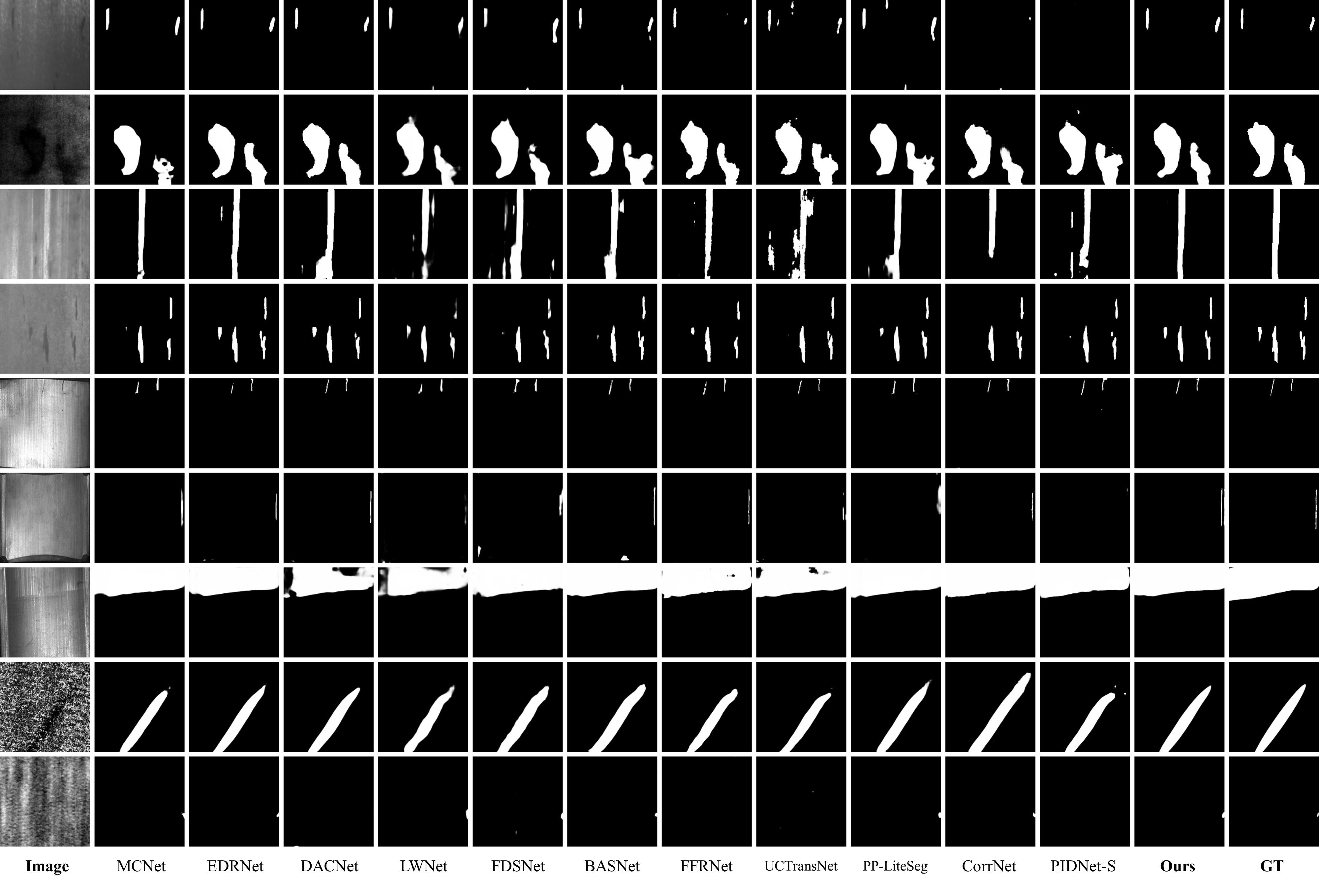}
	\caption{Visual Comparisons of different models on SD-saliency-900 ($\rm 1^{st} \sim 4^{th}$ rows), Magnetic tile ($\rm  5^{th} \sim 7^{th}$ rows), and DAGM 2007 ($\rm 8^{th}$ and $\rm 9^{th}$ rows), respectively.}
	\label{fig5}
\end{figure*}

\subsubsection{Quantitative Comparison} Table \ref{tab1} presents the quantitative comparison results of various models on accuracy ($MAE$, $S_m$, $F_{\beta}^{w}$ and $E_{\phi}$ ) and running efficiency (\#Params, FLOPs, and FPS).
%
%
%
It is observed that the proposed GCANet obtains the best trade-off between accuracy and efficiency.
Compared with the defect detection model DACNet\cite{Zhou2022DACNet}, the proposed method only decreases by 0.96\%, 0.62\% and 0.38\% on $F_{\beta}^{w}$, $S_m$ and $E_{\phi}$ on SD-saliency-900, but with $98\times$ fewer FLOPs, $53\times$ fewer parameters and $7\times$ faster speed. 
The proposed method is even superior to \cite{Zhou2022DACNet} in terms of $F_{\beta}^{w}$, $S_m$ and $F_{\beta}$ on Magnetic tile and DAGM 2007.
Although lightweight defect detection models such as FDSNet \cite{zhang2022fdsnet} have achieved faster speed with fewer parameters and FLOPs, they are much lower than other defect detection methods (e.g., MCNet, EDRNet, and DACNet) in performance. 
And the proposed method also achieves competitive results in defect scenes in comparison with other efficient models.
Meanwhile, the experimental results also show that most existing lightweight models perform poorly in defect scenes. This suggests that these lightweight methods are not appropriate for defect detection since defect objects are more complex than other objects. Compared with these methods, our lightweight method obtains better performance in defect scenes.

Fig. \ref{fig3} intuitively illustrates the trade-off comparison of various methods between accuracy (average $F_{\beta}^{w}$ of three datasets) and running efficiency (\#Param). The proposed method (the red dot) lies at the top-left corner, which obviously suggests that our method achieves competitive performance with fewer parameters.
In addition, Fig. \ref{fig4} indicates that the proposed GCANet (the red curve) also obtains competitive performance with respect to the PR and F-measure curves over three defect datasets. 

\begin{table}[!t]
\renewcommand\arraystretch{1.3}
\centering
\caption{Quantitative comparisons of different methods on SD-saliency-900 ($\rho=20\%$). }
\label{tab2}
\setlength{\tabcolsep}{1mm}{

\begin{tabular}{c|r|ccc}		
\hline
\multirow{2}*{No.}&\multirow{2}*{Methods} &\multicolumn{3}{c}{\tabincell{c}{SD-saliency-900 ($\rho=20\%$)}}  \\
\cline{3-5}
~&~ & $F_{\beta}^{w} \uparrow$ & $S_{\alpha} \uparrow$ &$E_{\phi} \uparrow$ \\
\hline
1&MCNet\cite{Zhang2021MCnet} 
&0.8128$\downarrow_{9.38}$&0.8573$\downarrow_{6.84}$&0.9144$\downarrow_{5.47}$

\\
2&EDRNet\cite{Song2020EDRNet} 
&0.9056$\downarrow_{1.69}$&0.9244$\downarrow_{1.31}$&0.9701$\downarrow_{0.53}$

\\
3&DACNet\cite{Zhou2022DACNet} 
&0.9087$\downarrow_{1.88}$&0.9238$\downarrow_{1.79}$&0.9679$\downarrow_{0.94}$

\\
4&LWNet\cite{Huang2020Compact}
&0.8166$\downarrow_{3.50}$&0.8688$\downarrow_{2.29}$&0.9376$\downarrow_{1.57}$

\\
5&FDSNet\cite{zhang2022fdsnet} 
&0.8199$\downarrow_{6.71}$&0.8635$\downarrow_{4.78}$&0.9401$\downarrow_{2.47}$

\\
\hline
6&BASNet\cite{Qin2019basnet} & 0.9014$\downarrow_{0.78}$& 0.9219$\downarrow_{0.57}$ & 0.9678 $\downarrow_{0.13}$

\\
7&FFRNet\cite{Wang2022Focus}& 0.7697$\downarrow_{14.75}$&0.8299$\downarrow_{10.54}$&0.8761$\downarrow_{9.70}$

\\
8&ACCoNet\cite{li2022adjacent}&  0.8349$\downarrow_{6.95}$&0.8687$\downarrow_{5.30}$&0.9283$\downarrow_{3.85}$

\\

9&PGNet \cite{Xie2022PGNet} 
& 0.8190$\downarrow_{8.97}$&0.8628$\downarrow_{6.68}$&0.9216$\downarrow_{4.88}$

\\
10&UCTransNet \cite{2022wangUCTransNet}
& 0.8539$\downarrow_{6.34}$&0.8845$\downarrow_{5.07}$&0.9389$\downarrow_{3.39}$

\\
\hline
11&HVPNet\cite{liu2021lightweight} & 0.8701$\downarrow_{1.85}$&0.9040$\downarrow_{1.49}$&0.9534$\downarrow_{0.65}$
\\
12&STDC2-Seg\cite{fan2021rethinking} 
&0.8680$\downarrow_{3.27}$&0.8962$\downarrow_{2.46}$&0.9526$\downarrow_{1.67}$

\\
13&SAMNet\cite{liu2021samnet}
&0.8462$\downarrow_{3.94}$&0.8866$\downarrow_{3.15}$&0.9426$\downarrow_{1.68}$

\\
14&AttaNet\cite{Song2021AttaNet} 
&0.8011$\downarrow_{9.21}$&0.8537$\downarrow_{6.52}$&0.9126$\downarrow_{5.62}$

\\
15&PP-LiteSeg\cite{peng2022pp} 
&0.8770$\downarrow_{2.51}$&0.9031$\downarrow_{2.13}$&0.9596$\downarrow_{1.07}$

\\
16&CorrNet\cite{li2022CorrNet}& 0.8892$\downarrow_{2.15}$&0.9102$\downarrow_{0.70}$&0.9578$\downarrow_{1.05}$\\
17&PIDNet-S\cite{xu2023pidnet}&0.7979$\downarrow_{11.1}$&0.8522$\downarrow_{7.87}$&0.8841$\downarrow_{8.13}$
\\
\hline
18&GCANet(Ours) &  0.9033$\downarrow_{1.46}$&0.9226$\downarrow_{1.29}$&0.9698$\downarrow_{0.37}$
\\
\hline
\end{tabular}}
\end{table}
To demonstrate the robustness of the proposed method against background interference,  we further compare the performance of various methods on the SD-saliency-900 with severe salt-and-pepper noise ($\rho=20\%$) added, similar to the previous works \cite{Song2020EDRNet, Zhou2022DACNet}. The experimental results in Table \ref{tab2} suggest that the proposed GCANet is less affected by interference noise on performance compared to other methods, which only reduces by 1.46\% 1.29\% and 0.37\% on $F_{\beta}^{w}$, $S_{\alpha}$ and $E_{\phi}$, respectively.

Overall, compared with other methods, the proposed method displays competitive performance on six evaluation metrics while maintaining lower computational overhead. This implies that the proposed GCANet is an efficient lightweight defect detection model.

\subsubsection{Visual Comparison} Fig. \ref{fig5} displays some detection results of various methods on three defect datasets. The visualization results show that the complexity of defects may result in some false or incomplete detection results.
For example, when there are some distractions in the background ($\rm 1^{st}$ and $ \rm 3^{rd}$ rows), some methods are prone to making false predictions, detecting them as defective regions. 
Some methods fail to detect some defect details, such as small defects ($\rm 4^{th}$ and $\rm 9^{th}$ rows) and fine scratches ($\rm 5^{th}$ and $\rm 6^{th}$ rows).
And for some low-contrast defects ($\rm 2^{nd}$, $\rm 7^{th}$, and $\rm 8^{th}$ rows), which are highly similar to the background, some methods cannot detect them accurately either.
In contrast, the proposed method obtains more precise detection results than other methods on those challenging defects, which demonstrates the effectiveness of GCANet in defect scenarios.

\subsection{Ablation Study}
To investigate the validity of various modules, the following ablation experiments are done on SD-saliency-900 and DAGM 2007, respectively.

\subsubsection{Ablation for Network Architecture} 
\begin{figure}[t]
	\centering
		\includegraphics[scale=0.32]{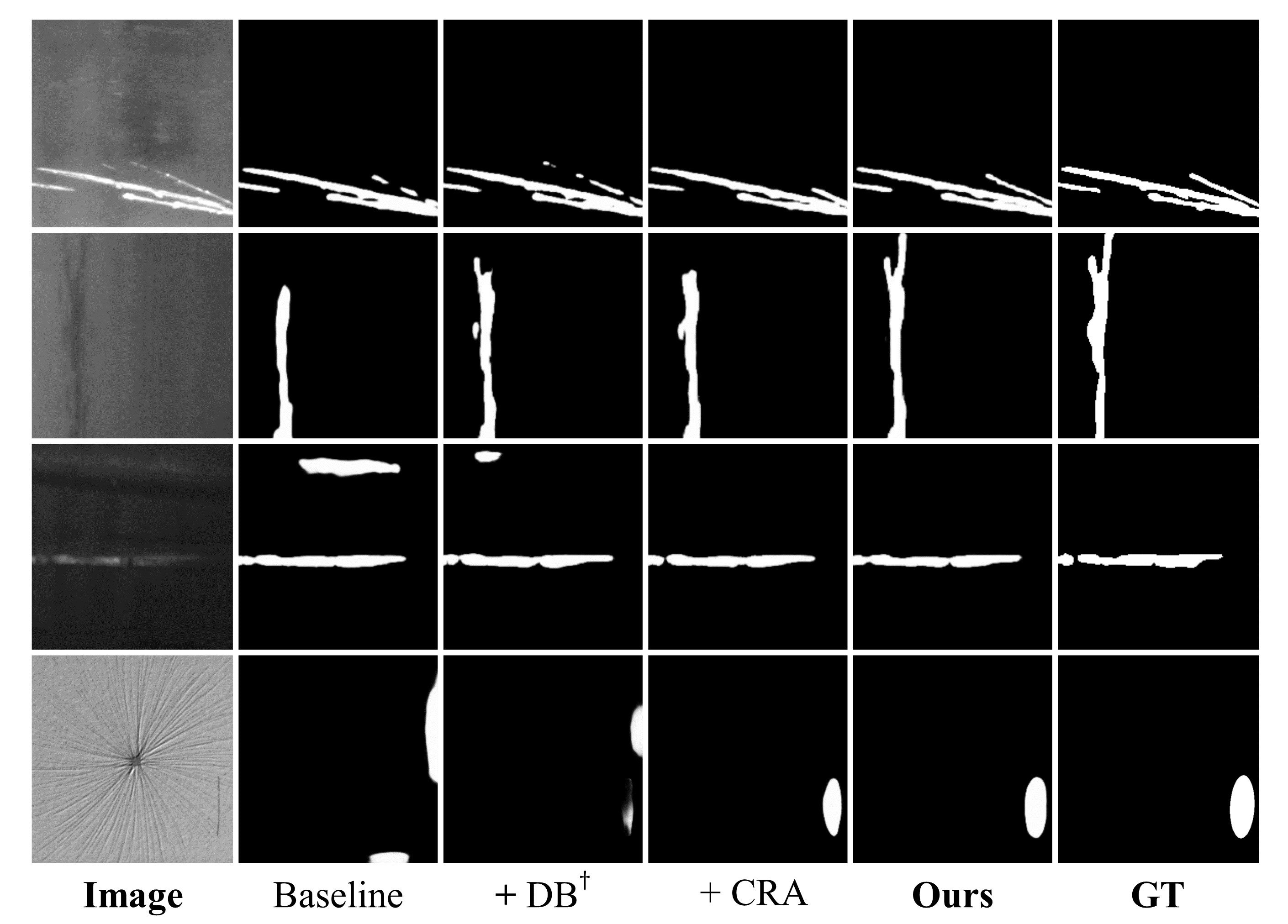}
	\caption{Visual comparison of models under different settings in Table \ref{tab3}.
	}
	\label{fig6}
\end{figure}

To prove the validity of each module in GCANet, we perform the ablation study for network architecture.
Specifically, the baseline adopts the same encoder and decoder as the lightweight SOD model SAMNet, except that the input is not downsampled in the first encoder stage.
Based on that, we introduce the lightweight transformer encoder after the last encoder stage, and then add different modules, respectively.
As shown in Table \ref{tab3}, the injection of global information in each decoder block greatly improves the performance of the baseline. But as illustrated in Fig. \ref{fig6}, the existence of background interference brings some wrong or incomplete detection results.
Similarly, the introduction of the CRA module also brings significant performance improvements but only incurs little computational overhead compared to the baseline. 
It can make the network focus more on defect objects, as illustrated in Fig. \ref{fig6}.
By simultaneously introducing CRA and injecting global information into each decoder block, the model can detect more accurate defect regions, yielding the best performance.
This indicates that it is effective for defect detection to strengthen feature representation and introduce global information in the network.

\begin{table}[!t]
\renewcommand\arraystretch{1.3}
\centering
\caption{The ablation study for the proposed GCANet. }
\label{tab3}
\setlength{\tabcolsep}{0.3mm}{
\begin{tabular}{r|c|c|ccc|ccc}		
\hline
\multirow{2}*{Settings} &\multirow{2}*{\tabincell{c}{\#Param\\(M)}}&\multirow{2}*{\tabincell{c}{FLOPs\\(G)}}&\multicolumn{3}{c|}{\tabincell{c}{SD-saliency-900}}&\multicolumn{3}{c}{\tabincell{c}{DAGM 2007}}\\
\cline{4-9}
~ &~ &~ &$\text{MAE}\downarrow$ & $F_{\beta}^{w}\uparrow$ & $S_{\alpha}\uparrow$&$\text{MAE}\downarrow$ & $F_{\beta}^{w}\uparrow$ & $S_{\alpha}\uparrow$   \\
\hline
Baseline&1.28&1.20&0.0168& 0.8994&0.9237&0.0055&0.8429&0.9060\\
+ DB$^{\dag}$ &1.74&1.37&0.0144& 0.9140&0.9311&0.0046&0.8560&0.9111\\
+ CRA &1.81&1.45&0.0147& 0.9112&0.9300&0.0049&0.8536&0.9095\\
GCANet(Ours)&1.84&1.45& 0.0135& 0.9179&0.9355&0.0044&0.8653&0.9151\\
\hline
\end{tabular}}
\begin{tablenotes}   
        \footnotesize               
        \item[1] DB$^{\dag}$ represents that the DB takes the inputs as global features besides high-level and low-level features.
      \end{tablenotes}  
\end{table}

\begin{table}[!t]
\renewcommand\arraystretch{1.3}
\centering
\caption{Quantitative comparison for DSA and other modules. The "MSA-8" denotes the multi-head self-attention with 8 heads.}
\label{tab4}
\setlength{\tabcolsep}{0.32mm}{
\begin{tabular}{r|c|c|ccc|ccc}		
\hline
\multirow{2}*{Settings} &\multirow{2}*{\tabincell{c}{\#Param\\(M)}}&\multirow{2}*{\tabincell{c}{FLOPs\\(G)}}&\multicolumn{3}{c|}{\tabincell{c}{SD-saliency-900}}&\multicolumn{3}{c}{\tabincell{c}{DAGM 2007}}\\
\cline{4-9}
~ &~ &~ &$\text{MAE}\downarrow$ & $F_{\beta}^{w}\uparrow$ & $S_{\alpha}\uparrow$&$\text{MAE}\downarrow$ & $F_{\beta}^{w}\uparrow$ & $S_{\alpha}\uparrow$  \\
\hline
w PPM \cite{zhao2017pyramid} &1.47&1.40&0.0148& 0.9107&0.9289&0.0047&0.8518&0.9077\\
w MSA-8 \cite{vaswani2017attention} &1.84&1.50 &0.0140& 0.9154&0.9328&0.0046&0.8623&0.9137\\
GCANet(Ours)&1.84&1.45& 0.0135& 0.9179&0.9355&0.0044&0.8653&0.9151\\
\hline
\end{tabular}}
 
\end{table}

\subsubsection{Effectiveness of DSA} To prove the effectiveness of the proposed DSA module, it is compared with traditional MSA \cite{vaswani2017attention}, as shown in Table \ref{tab4}. Specifically, we replace DSA in each transformer block with MSA (denoted as w MSA). It is found that the proposed DSA reduces FOLPs by 50M but obtains better performance compared with MSA.
Furthermore, we compare the transformer encoder with PPM \cite{zhao2017pyramid} in Table \ref{tab4}, where we replace the transformer encoder with PPM (denoted as w PPM). 
The experimental results show that the transformer encoder obtains better performance than the PPM in the network. The main reason is that pooling operations damage local details, resulting in poor performance.
The experimental results demonstrate the effectiveness of DSA. 

\section{Conclusion}
In this paper, we propose a Global Context Aggregation Network (GCANet) for lightweight saliency detection of surface defects. 
GCANet introduces a novel transformer encoder to learn global information, remedying the limitations of CNNs lacking global information. In each transformer block, Depth-wise Self-Attention (DSA) module produces global weights by element-wise interaction between features, implicitly modeling global dependencies.
In addition, Channel Reference Attention (CRA) module is embedded before each decoder block for feature enhancement. CRA utilizes the interaction of cross-layer features to suppress background interference and mine important defect details.
The extensive experiments on three defect datasets demonstrate that the proposed lightweight method achieves promising performance with low computational costs compared to 17 other state-of-the-art methods.
\bibliographystyle{ieeetr}
\bibliography{GCANet}

\begin{thebibliography}{10}

\bibitem{hanmandlu2015detection}
M.~Hanmandlu, D.~Choudhury, and S.~Dash, ``Detection of defects in fabrics
  using topothesy fractal dimension features,'' {\em Signal, Image and Video
  Processing}, vol.~9, no.~7, pp.~1521--1530, 2015.

\bibitem{SONG2020LBP}
G.~Song, K.~Song, and Y.~Yan, ``Saliency detection for strip steel surface
  defects using multiple constraints and improved texture features,'' {\em
  Optics and Lasers in Engineering}, vol.~128, p.~106000, 2020.

\bibitem{Jiang2020Gabor}
W.~Jiang, T.~Li, and B.~Shi, ``Classification of surface defects based on
  improved gabor filter,'' in {\em Proceedings of the International Conference
  on Control, Robotics and Cybernetics}, pp.~151--155, 2020.

\bibitem{Berwo2021Canny}
M.~A. Berwo, Y.~Fang, J.~Mahmood, and E.~A. Retta, ``Automotive engine cylinder
  head crack detection: Canny edge detection with morphological dilation,'' in
  {\em Proceedings of Asia-Pacific Signal and Information Processing
  Association Annual Summit and Conference}, pp.~1519--1527, 2021.

\bibitem{huang2020surface}
Y.~Huang, C.~Qiu, and K.~Yuan, ``Surface defect saliency of magnetic tile,''
  {\em The Visual Computer}, vol.~36, no.~1, pp.~85--96, 2020.

\bibitem{li2021fast}
K.~Li, B.~Wang, Y.~Tian, and Z.~Qi, ``Fast and accurate road crack detection
  based on adaptive cost-sensitive loss function,'' {\em IEEE Transactions on
  Cybernetics}, vol.~53, no.~2, pp.~1051--1062, 2023.

\bibitem{fan2022rethinking}
R.~Fan, U.~Ozgunalp, Y.~Wang, M.~Liu, and I.~Pitas, ``Rethinking road surface
  3-d reconstruction and pothole detection: From perspective transformation to
  disparity map segmentation,'' {\em IEEE Transactions on Cybernetics},
  vol.~52, no.~7, pp.~5799--5808, 2022.

\bibitem{niu2021Unsupervised}
M.~Niu, K.~Song, L.~Huang, Q.~Wang, Y.~Yan, and Q.~Meng, ``Unsupervised
  saliency detection of rail surface defects using stereoscopic images,'' {\em
  IEEE Transactions on Industrial Informatics}, vol.~17, no.~3, pp.~2271--2281,
  2021.

\bibitem{Zhang2021MCnet}
D.~Zhang, K.~Song, J.~Xu, Y.~He, M.~Niu, and Y.~Yan, ``Mcnet: Multiple context
  information segmentation network of no-service rail surface defects,'' {\em
  IEEE Transactions on Instrumentation and Measurement}, vol.~70, pp.~1--9,
  2021.

\bibitem{Song2020EDRNet}
G.~Song, K.~Song, and Y.~Yan, ``Edrnet: Encoder–decoder residual network for
  salient object detection of strip steel surface defects,'' {\em IEEE
  Transactions on Instrumentation and Measurement}, vol.~69, no.~12,
  pp.~9709--9719, 2020.

\bibitem{Zhou2022DACNet}
X.~Zhou, H.~Fang, Z.~Liu, B.~Zheng, Y.~Sun, J.~Zhang, and C.~Yan, ``Dense
  attention-guided cascaded network for salient object detection of strip steel
  surface defects,'' {\em IEEE Transactions on Instrumentation and
  Measurement}, vol.~71, pp.~1--14, 2022.

\bibitem{wan2023lfrnet}
B.~Wan, X.~Zhou, B.~Zheng, H.~Yin, Z.~Zhu, H.~Wang, Y.~Sun, J.~Zhang, and
  C.~Yan, ``Lfrnet: Localizing, focus, and refinement network for salient
  object detection of surface defects,'' {\em IEEE Transactions on
  Instrumentation and Measurement}, vol.~72, pp.~1--12, 2023.

\bibitem{bovzivc2021mixed}
J.~Bo{\v{z}}i{\v{c}}, D.~Tabernik, and D.~Sko{\v{c}}aj, ``Mixed supervision for
  surface-defect detection: From weakly to fully supervised learning,'' {\em
  Computers in Industry}, vol.~129, p.~103459, 2021.

\bibitem{yang2022Semic}
Y.~Yang and M.~Sun, ``Semiconductor defect detection by hybrid
  classical-quantum deep learning,'' in {\em Proceedings of the {IEEE/CVF}
  Conference on Computer Vision and Pattern Recognition}, pp.~2313--2322, 2022.

\bibitem{Gaurab2021Interleaved}
G.~Bhattacharya, B.~Mandal, and N.~B. Puhan, ``Interleaved deep artifacts-aware
  attention mechanism for concrete structural defect classification,'' {\em
  IEEE Transactions on Image Processing}, vol.~30, pp.~6957--6969, 2021.

\bibitem{su2021baf}
B.~Su, H.~Chen, and Z.~Zhou, ``Baf-detector: An efficient cnn-based detector
  for photovoltaic cell defect detection,'' {\em IEEE Transactions on
  Industrial Electronics}, vol.~69, no.~3, pp.~3161--3171, 2021.

\bibitem{Cui2021SDD}
L.~Cui, X.~Jiang, M.~Xu, W.~Li, P.~Lv, and B.~Zhou, ``Sddnet: A fast and
  accurate network for surface defect detection,'' {\em IEEE Transactions on
  Instrumentation and Measurement}, vol.~70, pp.~1--13, 2021.

\bibitem{Ni2022ANet}
X.~Ni, Z.~Ma, J.~Liu, B.~Shi, and H.~Liu, ``Attention network for rail surface
  defect detection via consistency of intersection-over-union(iou)-guided
  center-point estimation,'' {\em IEEE Transactions on Industrial Informatics},
  vol.~18, no.~3, pp.~1694--1705, 2022.

\bibitem{Sampath2023Attention}
V.~Sampath, I.~Maurtua, J.~J.~A. Martín, A.~Rivera, J.~Molina, and
  A.~Gutierrez, ``Attention-guided multitask learning for surface defect
  identification,'' {\em IEEE Transactions on Industrial Informatics}, vol.~19,
  no.~9, pp.~9713--9721, 2023.

\bibitem{wang2023rern}
C.~Wang, H.~Chen, and S.~Zhao, ``Rern: Rich edge features refinement detection
  network for polycrystalline solar cell defect segmentation,'' {\em IEEE
  Transactions on Industrial Informatics}, pp.~1--12, 2023.

\bibitem{zhang2022fdsnet}
J.~Zhang, R.~Ding, M.~Ban, and T.~Guo, ``Fdsnet: An accurate real-time surface
  defect segmentation network,'' in {\em Proceedings of the IEEE International
  Conference on Acoustics, Speech and Signal Processing}, pp.~3803--3807, 2022.

\bibitem{fan2021rethinking}
M.~Fan, S.~Lai, J.~Huang, X.~Wei, Z.~Chai, J.~Luo, and X.~Wei, ``Rethinking
  bisenet for real-time semantic segmentation,'' in {\em Proceedings of the
  IEEE/CVF Conference on Computer Vision and Pattern Recognition},
  pp.~9716--9725, 2021.

\bibitem{peng2022pp}
J.~Peng, Y.~Liu, S.~Tang, Y.~Hao, L.~Chu, G.~Chen, Z.~Wu, Z.~Chen, Z.~Yu,
  Y.~Du, {\em et~al.}, ``Pp-liteseg: A superior real-time semantic segmentation
  model,'' {\em arXiv preprint arXiv:2204.02681}, 2022.

\bibitem{xu2023pidnet}
J.~Xu, Z.~Xiong, and S.~P. Bhattacharyya, ``Pidnet: A real-time semantic
  segmentation network inspired by pid controllers,'' in {\em Proceedings of
  the IEEE/CVF Conference on Computer Vision and Pattern Recognition},
  pp.~19529--19539, 2023.

\bibitem{liu2021lightweight}
Y.~Liu, Y.-C. Gu, X.-Y. Zhang, W.~Wang, and M.-M. Cheng, ``Lightweight salient
  object detection via hierarchical visual perception learning,'' {\em IEEE
  Transactions on Cybernetics}, vol.~51, no.~9, pp.~4439--4449, 2021.

\bibitem{liu2021samnet}
Y.~Liu, X.-Y. Zhang, J.-W. Bian, L.~Zhang, and M.-M. Cheng, ``Samnet:
  Stereoscopically attentive multi-scale network for lightweight salient object
  detection,'' {\em IEEE Transactions on Image Processing}, vol.~30,
  pp.~3804--3814, 2021.

\bibitem{li2021Depthwise}
H.~Li, G.~Li, B.~Yang, G.~Chen, L.~Lin, and Y.~Yu, ``Depthwise nonlocal module
  for fast salient object detection using a single thread,'' {\em IEEE
  Transactions on Cybernetics}, vol.~51, no.~12, pp.~6188--6199, 2021.

\bibitem{peng2021conformer}
Z.~Peng, W.~Huang, S.~Gu, L.~Xie, Y.~Wang, J.~Jiao, and Q.~Ye, ``Conformer:
  Local features coupling global representations for visual recognition,'' in
  {\em Proceedings of the IEEE/CVF International Conference on Computer
  Vision}, pp.~367--376, 2021.

\bibitem{Xie2022PGNet}
C.~Xie, C.~Xia, M.~Ma, Z.~Zhao, X.~Chen, and J.~Li, ``Pyramid grafting network
  for one-stage high resolution saliency detection,'' in {\em Proceedings of
  the IEEE Conference on Computer Vision and Pattern Recognition},
  pp.~11717--11726, 2022.

\bibitem{Liu21PamiPoolNet}
J.-J. Liu, Q.~Hou, Z.-A. Liu, and M.-M. Cheng, ``Poolnet+: Exploring the
  potential of pooling for salient object detection,'' {\em IEEE Transactions
  on Pattern Analysis and Machine Intelligence}, vol.~45, no.~1, pp.~887--904,
  2023.

\bibitem{zhao2017pyramid}
H.~Zhao, J.~Shi, X.~Qi, X.~Wang, and J.~Jia, ``Pyramid scene parsing network,''
  in {\em Proceedings of the IEEE conference on computer vision and pattern
  recognition}, pp.~2881--2890, 2017.

\bibitem{vaswani2017attention}
A.~Vaswani, N.~Shazeer, N.~Parmar, J.~Uszkoreit, L.~Jones, A.~N. Gomez,
  {\L}.~Kaiser, and I.~Polosukhin, ``Attention is all you need,'' {\em Advances
  in neural information processing systems}, vol.~30, 2017.

\bibitem{Huang2020Compact}
Y.~Huang, C.~Qiu, X.~Wang, S.~Wang, and K.~Yuan, ``A compact convolutional
  neural network for surface defect inspection,'' {\em Sensors}, vol.~20,
  no.~7, p.~1974, 2020.

\bibitem{Poudel2019FastSCNNFS}
R.~P.~K. Poudel, S.~Liwicki, and R.~Cipolla, ``Fast-scnn: Fast semantic
  segmentation network,'' in {\em Proceedings of British Machine Vision
  Conference}, 2019.

\bibitem{li2022CorrNet}
G.~Li, Z.~Liu, Z.~Bai, W.~Lin, and H.~Ling, ``Lightweight salient object
  detection in optical remote sensing images via feature correlation,'' {\em
  IEEE Transactions on Geoscience and Remote Sensing}, vol.~60, pp.~1--12,
  2022.

\bibitem{Hou2017deeply}
Q.~Hou, M.-M. Cheng, X.~Hu, A.~Borji, Z.~Tu, and P.~H. Torr, ``Deeply
  supervised salient object detection with short connections,'' in {\em
  Proceedings of the IEEE Conference on Computer Vision and Pattern
  Recognition}, pp.~3203--3212, 2017.

\bibitem{BoerKMR05}
P.~de~Boer, D.~P. Kroese, S.~Mannor, and R.~Y. Rubinstein, ``A tutorial on the
  cross-entropy method,'' {\em Annals of Operations Research}, vol.~134, no.~1,
  pp.~19--67, 2005.

\bibitem{Rahman2016Optimizing}
M.~A. Rahman and Y.~Wang, ``Optimizing intersection-over-union in deep neural
  networks for image segmentation,'' in {\em Proceedings of the International
  Symposium on Visual Computing}, vol.~10072, pp.~234--244, 2016.

\bibitem{Wang2003structural}
Z.~Wang, E.~Simoncelli, and A.~Bovik, ``Multiscale structural similarity for
  image quality assessment,'' in {\em Proceedings of the Thrity-Seventh
  Asilomar Conference on Signals, Systems \& Computers}, vol.~2,
  pp.~1398--1402, 2003.

\bibitem{wang2023defect}
J.~Wang, G.~Xu, F.~Yan, J.~Wang, and Z.~Wang, ``Defect transformer: An
  efficient hybrid transformer architecture for surface defect detection,''
  {\em Measurement}, vol.~211, p.~112614, 2023.

\bibitem{wieler2007weakly}
M.~Wieler and T.~Hahn, ``Weakly supervised learning for industrial optical
  inspection,'' in {\em DAGM symposium in}, vol.~6, 2007.

\bibitem{borji2019salient}
A.~Borji, M.~Cheng, Q.~Hou, H.~Jiang, and J.~Li, ``Salient object detection: A
  survey,'' {\em Computational Visual Media}, vol.~5, no.~2, pp.~117--150,
  2019.

\bibitem{achanta2009frequency}
R.~Achanta, S.~Hemami, F.~Estrada, and S.~S{\"u}sstrunk, ``Frequency-tuned
  salient region detection,'' in {\em Proceedings of the IEEE Conference on
  Computer Vision and Pattern Recognition}, pp.~1597--1604, 2009.

\bibitem{perazzi2012saliency}
F.~Perazzi, P.~Kr{\"a}henb{\"u}hl, Y.~Pritch, and A.~Hornung, ``Saliency
  filters: Contrast based filtering for salient region detection,'' in {\em
  Proceedings of the IEEE Conference on Computer Vision and Pattern
  Recognition}, pp.~733--740, 2012.

\bibitem{margolin2014evaluate}
R.~Margolin, L.~Zelnik-Manor, and A.~Tal, ``How to evaluate foreground maps,''
  in {\em Proceedings of the IEEE Conference on Computer Vision and Pattern
  Recognition}, pp.~248--255, 2014.

\bibitem{fan2017structure}
D.-P. Fan, M.-M. Cheng, Y.~Liu, T.~Li, and A.~Borji, ``Structure-measure: A new
  way to evaluate foreground maps,'' in {\em Proceedings of the IEEE
  International Conference on Computer Vision}, pp.~4548--4557, 2017.

\bibitem{EM}
D.-P. Fan, C.~Gong, Y.~Cao, B.~Ren, M.-M. Cheng, and A.~Borji,
  ``Enhanced-alignment measure for binary foreground map evaluation,'' in {\em
  Proceedings of the International Joint Conference on Artificial
  Intelligence}, pp.~698--704, 2018.

\bibitem{Qin2019basnet}
X.~Qin, Z.~Zhang, C.~Huang, C.~Gao, M.~Dehghan, and M.~Jagersand, ``Basnet:
  Boundary-aware salient object detection,'' in {\em Proceedings of the
  IEEE/CVF Conference on Computer Vision and Pattern Recognition},
  pp.~7471--7481, 2019.

\bibitem{Wang2022Focus}
R.~Wang, C.~Ji, Y.~Zhang, and Y.~Li, ``Focus, fusion, and rectify:
  Context-aware learning for {COVID-19} lung infection segmentation,'' {\em
  IEEE Transactions on Neural Networks and Learning Systems}, vol.~33, no.~1,
  pp.~12--24, 2022.

\bibitem{li2022adjacent}
G.~Li, Z.~Liu, D.~Zeng, W.~Lin, and H.~Ling, ``Adjacent context coordination
  network for salient object detection in optical remote sensing images,'' {\em
  IEEE Transactions on Cybernetics}, vol.~52, no.~7, pp.~5799--5808, 2022.

\bibitem{2022wangUCTransNet}
H.~Wang, P.~Cao, J.~Wang, and O.~R. Zaiane, ``Uctransnet: rethinking the skip
  connections in u-net from a channel-wise perspective with transformer,'' in
  {\em Proceedings of the AAAI conference on artificial intelligence},
  pp.~2441--2449, 2022.

\bibitem{Song2021AttaNet}
Q.~Song, K.~Mei, and R.~Huang, ``Attanet: Attention-augmented network for fast
  and accurate scene parsing,'' in {\em {Proceedings of the AAAI Conference on
  Artificial Intelligence}}, pp.~2567--2575, 2021.

\end{thebibliography}
\end{document}